\documentclass[10pt,twocolumn,letterpaper]{article}

\usepackage[pagenumbers]{cvpr} 

\usepackage{times}  
\usepackage{helvet}  
\usepackage{bbding}
\usepackage{amssymb}
\usepackage{courier}  
\usepackage[hyphens]{url}  
\usepackage{graphicx} 
\usepackage{wasysym} 
\usepackage{natbib}  
\usepackage{caption} 
\usepackage{algorithm}
\usepackage{algorithmicx}
\usepackage{algpseudocode}
\usepackage{booktabs}
\usepackage{multirow}

\usepackage{xcolor}
\usepackage{colortbl}
\usepackage{tikz}
\usepackage{enumitem}
\newcommand{\circled}[1]{\tikz[baseline=(char.base)]{
    \node[shape=circle, draw, inner sep=0.5pt,     
          minimum size=0.9em]                      
    (char) {#1};}}

\newlist{circledlist}{enumerate}{1}
\setlist[circledlist]{label=\protect\circled{\arabic*}}
\usepackage{newfloat}
\usepackage{listings}
\usepackage{booktabs}
\usepackage{multirow} 
\usepackage{float}

\usepackage{graphicx}
\usepackage{amsmath,amsthm,amssymb,amsfonts,xspace}

\usepackage{color}

\usepackage{colortbl}

\definecolor{cvprblue}{rgb}{0.21,0.49,0.74}
\definecolor{mycolor}{rgb}{0.9,0.9,0.9}
\usepackage[pagebackref,breaklinks,colorlinks,allcolors=cvprblue]{hyperref}

\def\paperID{19089} 
\def\confName{CVPR}
\def\confYear{2026}

\title{UAV-MM3D: A Large-Scale Synthetic Benchmark for 3D Perception of Unmanned Aerial Vehicles with Multi-Modal Data}

\author{Longkun Zou\\
Institution1\\
Institution1 address\\
{\tt\small firstauthor@i1.org}
\and
Second Author\\
Institution2\\
First line of institution2 address\\
{\tt\small secondauthor@i2.org}
}

\author{
Longkun Zou$^{1,*}$ \quad
Jiale Wang$^{2,*,\dagger}$ \quad
Rongqin Liang$^{1}$ \quad
Hai Wu$^{1}$ \quad
Ke Chen$^{1,}$\textsuperscript{\Envelope} \quad
Yaowei Wang$^{1}$\\[2mm]
$^{1}$ Pengcheng Laboratory \\
$^{2}$ University of Southern California 
}

\begin{document}



\maketitle
\begingroup
    \renewcommand\thefootnote{}
    \footnote{$^{*}$ Equal contribution \quad {\Envelope} Corresponding author}\\
    \footnote{$^{\dagger}$ The work was done during internship at the Pengcheng Laboratory.}
    \addtocounter{footnote}{-1}
\endgroup

\begin{abstract}
Accurate perception of UAVs in complex low-altitude environments is critical for airspace security and related intelligent systems. Developing reliable solutions requires large-scale, accurately annotated, and multimodal data. However, real-world UAV data collection faces inherent constraints due to airspace regulations, privacy concerns, and environmental variability, while manual annotation of 3D poses and cross-modal correspondences is time-consuming and costly. To overcome these challenges, we introduce UAV-MM3D, a high-fidelity multimodal synthetic dataset for low-altitude UAV perception and motion understanding. It comprises 400K synchronized frames across diverse scenes (urban areas, suburbs, forests, coastal regions) and weather conditions (clear, cloudy, rainy, foggy), featuring multiple UAV models (micro, small, medium-sized) and five modalities — RGB, IR, LiDAR, Radar, and DVS (Dynamic Vision Sensor). Each frame provides 2D/3D bounding boxes, 6-DoF poses, and instance-level annotations, enabling core tasks related to UAVs such as 3D detection, pose estimation, target tracking, and short-term trajectory forecasting. We further propose LGFusionNet, a LiDAR-guided multimodal fusion baseline, and a dedicated UAV trajectory prediction baseline to facilitate benchmarking. With its controllable simulation environment, comprehensive scenario coverage, and rich annotations, UAV3D offers a public benchmark for advancing 3D perception of UAVs.
\end{abstract}
    
\section{Introduction}

The rapid proliferation of unmanned aerial vehicles (UAVs) in consumer, commercial, and industrial applications has significantly expanded low-altitude airspace activities. However, this growth also introduces severe challenges to low-altitude security, critical infrastructure protection, and public privacy~\cite{Valavanis2015,UAVthreat2020}. In complex low-altitude environments—characterized by diverse urban, suburban, forest, and coastal scenes, adverse weather conditions (e.g., rain, fog, strong sunlight), and fast-moving, small-sized aerial targets—accurate and real-time UAV perception (including detection, 3D pose estimation, tracking, and trajectory forecasting) plays a central role in low-altitude airspace management and security systems~\cite{AntiUAVSurvey2021,Guo2022FusionReview}. Despite substantial progress in aerial perception, the development of robust and generalizable UAV perception models is fundamentally constrained by the lack of large-scale, multimodal datasets.

Existing UAV-related datasets exhibit several limitations that hinder further advancements. \textbf{First}, real-world UAV data collection is inherently constrained by strict airspace regulations, limited access to sensitive regions, and strong legal and privacy concerns regarding aerial imaging~\cite{Regulation2020}. Meanwhile, environmental stochasticity (e.g., abrupt weather changes) makes it difficult to systematically capture diverse real-world conditions. \textbf{Second}, multimodal sensing remains insufficient: most datasets focus on RGB imagery or RGB–thermal pairs~\cite{Anti-UAV-RGBT,DroneRGBT}, and only a few include LiDAR~\cite{MMDrone,AirSim2017}. Crucial modalities such as radar—important for long-range detection under occlusion—and Dynamic Vision Sensors (DVS)—essential for capturing high-speed motion—are rarely integrated, limiting robustness in complex scenarios~\cite{EventCameraSurvey2019,RadarSurvey2021}. \textbf{Third}, annotation richness is limited: most datasets provide only 2D bounding boxes without accurate 3D bounding boxes, 6-degree-of-freedom (6-DoF) pose annotations, or instance-level identity information~\cite{Anti-UAV600}. This restricts the development of multi-task methods for 3D detection, pose estimation, target tracking, and trajectory prediction. \textbf{Finally}, benchmark baselines for multimodal UAV perception are scarce, making it difficult to evaluate cross-modal alignment strategies and integrated perception models.

To overcome these limitations, we introduce \textbf{UAV-MM3D}, a large-scale, high-fidelity multimodal synthetic dataset tailored for low-altitude UAV perception and airborne motion understanding. Compared with real-world collections, simulation enables scalable, safe, and controllable data generation with comprehensive coverage of environmental conditions, sensor configurations, and UAV dynamics~\cite{SynthCity2019,DeepGTAV,Sim4CV}. UAV-MM3D incorporates five complementary modalities—RGB, infrared (IR), LiDAR, radar, and DVS—captured synchronously to leverage their complementary strengths: LiDAR provides accurate 3D geometric priors; radar offers robust long-range sensing under occlusion or adverse weather; DVS captures high-speed motion without motion blur; and RGB/IR ensure reliable perception across day–night transitions. The dataset contains 400,000 multi-sensor frames across diverse scenes, UAV types (micro, small, medium), weather conditions, and flight behaviors. Each frame is annotated with 2D and 3D bounding boxes, 6-DoF poses, instance IDs, and object categories, enabling research on 2D/3D detection, 6-DoF pose estimation, target tracking, and short-term trajectory prediction.

To further support benchmarking and research, we introduce two dedicated baseline models: \textbf{LGFusionNet}, a LiDAR-guided multimodal fusion network designed to align heterogeneous sensor features using LiDAR geometric priors; and a lightweight trajectory prediction baseline tailored to high-speed UAV motion patterns. Together with the dataset, these baselines provide a unified evaluation framework to standardize low-altitude UAV perception research.
The main contributions of this work are summarized as follows:
\begin{enumerate}
\item We present \textbf{UAV-MM3D}, a large-scale multimodal synthetic dataset containing 400K synchronized frames across five sensing modalities with comprehensive annotations. It fills the current gap in multimodal, multi-scenario, and multi-task UAV datasets.
\item UAV-MM3D provides diverse environments, weather conditions, UAV models, and precise 6-DoF annotations, supporting four essential low-altitude UAV perception tasks: 2D/3D detection, 6-DoF pose estimation, multi-object tracking, and trajectory prediction.
\item We introduce \textbf{LGFusionNet}, a LiDAR-guided multimodal fusion baseline, along with a trajectory prediction baseline, offering standardized benchmarks for evaluating multimodal UAV perception algorithms.
\end{enumerate}

\begin{table*}[t]
    \centering
    \setlength{\tabcolsep}{6pt}
    \resizebox{1.\linewidth}{!}{
        \begin{tabular}{l|c|c|c|c|c|c|c|c|c}
            \toprule
            Dataset & Modalities & 3D Boxes & 2D Boxes & Indoor/Outdoor & Class & Frames & Scenes & Day \& Weather & Data Type \\
            \midrule
            Anti-UAV-RGBT~\cite{Anti-UAV-RGBT} & RGB, IR & \XSolidBrush & \Checkmark & Outdoor & 6 & $\sim$297,000 & 3 & 2 & Real \\
            DUT Anti-UAV~\cite{DUTAntiUAV} & RGB, IR & \XSolidBrush & \Checkmark & Outdoor & \textbf{35} & $\sim$10,000 & – & 2 & Real \\
            Anti-UAV410~\cite{Anti-UAV410} & IR & \XSolidBrush & \Checkmark & Outdoor & 1 & 150,000  & 6 & 4 & Real \\ 
            MMAUD~\cite{Mmaud} & RGB, Audio, Radar, LiDAR & \XSolidBrush & \Checkmark & Outdoor & 5 & $\sim$50,000 & 1 & 1 & Real \\
            M3D~\cite{M3D} & RGB, Radar, LiDAR & \XSolidBrush & \Checkmark & Outdoor & 10 & 83,999 & 8 & 1 & Real/Sim \\
            DrIFT~\cite{DrIFT} & RGB, Radar & \XSolidBrush & \Checkmark & Outdoor & 7 & 47,991 & 3 & 4 & Real/Sim \\
            \midrule
            MAV6D~\cite{MAV6D} & RGB & \Checkmark & \XSolidBrush & Indoor & 2 & 33,489 & 1 & 1 & Real \\
            \rowcolor{gray!10}
            UAV-MM3D (Ours) & \textbf{RGB, IR, DVS, Radar, LiDAR} & \Checkmark & \Checkmark & Outdoor & 7 & \textbf{400,000} & \textbf{8} & \textbf{8} & Sim \\
            \bottomrule
        \end{tabular}
    }
    \caption{Comparison of recent low-altitude aerial (LAA) / anti-UAV datasets. Modalities: which sensors are included.}
    \label{dataset_table}
\end{table*}

\section{Related Work}

\textbf{Multimodal UAV datasets.}
The increasing deployment of UAVs in safety-critical low-altitude scenarios has encouraged the development of multimodal UAV datasets that combine heterogeneous sensing modalities. Existing datasets such as Anti-UAV-RGBT~\cite{Anti-UAV-RGBT}, Anti-UAV600~\cite{Anti-UAV600}, and CHASE\_DB~\cite{CHASEDB} provide RGB–thermal pairs for robust tracking or detection, especially under illumination changes. DroneRGBT~\cite{DroneRGBT} extends this direction by offering aligned RGB–thermal sequences for UAV tracking. MMDrone~\cite{MMDrone} introduces RGB–event modality pairing to address motion blur and high-speed flight. However, these datasets primarily focus on single-object tracking and offer only 2D annotations. In contrast, our proposed dataset offers aligned multimodal streams with complete 3D annotations suitable for 3D detection, 6-DoF pose estimation, and trajectory prediction of multiple aerial targets.

\textbf{Multimodal small object detection.}
Small object detection benefits greatly from multimodal fusion, especially under adverse conditions. Datasets such as KAIST Multispectral~\cite{KAIST}, and LLVIP~\cite{LLVIP} enable RGB–thermal fusion for small pedestrian detection under low-light settings. The DENSE dataset~\cite{DENSE} provides a rich combination of lidar, RGB, radar, and thermal sensors for all-weather perception, inspiring robust small-object perception via cross-modality complementarity. Similarly, RODNet~\cite{RODNet} introduces radar–camera fusion for detecting small and heavily occluded road objects. Nonetheless, most existing datasets are limited to ground-level scenes and lack low-altitude aerial targets. Our dataset fills this gap by offering multimodal imagery and synchronized 3D annotations for small flying objects, enabling more generalizable research on multimodal small-object detection in the sky.

\textbf{Multimodal autonomous driving perception datasets.}
Multimodal perception in autonomous driving has been extensively studied with datasets such as nuScenes~\cite{nuscenes}, Waymo~\cite{waymo}, and KITTI~\cite{kitti}, which provide rich sensor suites including cameras, lidar, and radar. Recent datasets such as Argoverse 2~\cite{Argoverse2} and ONCE~\cite{ONCE} scale up real-world diversity and offer dense 3D annotations for autonomous systems. Although these datasets demonstrate the effectiveness of multimodal fusion for 3D detection, tracking, and 6-DoF reasoning, they are constrained to road scenes and ground vehicles. The sensing geometry and target characteristics differ fundamentally from low-altitude aerial objects such as drones. Our dataset extends multimodal 3D perception research from road-plane targets to airborne objects with high maneuverability, richer orientation changes, and larger pose degrees of freedom.

\textbf{6-DoF pose estimation datasets.}
6-DoF object pose estimation has predominantly focused on static, ground-based objects with datasets such as LineMOD~\cite{LineMOD}, YCB-Video~\cite{YCB-Video}, and T-LESS~\cite{T-LESS}. Objectron~\cite{Objectron} and DexYCB~\cite{DexYCB} further provide category-level annotations and hand–object interactions. For aerial vehicles, MAV6D~\cite{MAV6D} is the most relevant dataset, offering 6-DoF pose labels for micro aerial vehicles. However, it is limited to small indoor environments and offers relatively small data volume. Our UAV-MMSim dataset provides large-scale outdoor multimodal data with accurate 3D bounding boxes and full 6-DoF poses, enabling more realistic studies in UAV 3D detection and tracking.

\section{The UAV-MM3D Dataset}
To support research on multimodal UAV and anti-UAV perception, we summarize representative UAV-related datasets in Table~\ref{dataset_table}. Existing datasets typically focus on limited sensing modalities, provide only 2D annotations, or lack sufficient environmental and viewpoint diversity. In contrast, our UAV-MM3D dataset is designed to deliver richer multimodal inputs, high-quality 3D annotations, and broader scenario coverage, providing a more comprehensive foundation for developing robust UAV perception and counter-UAV algorithms.

\subsection{Data Collection}

Despite the availability of various real-world UAV datasets, they typically fail to capture the full spectrum of environmental and weather diversity required for robust perception. To overcome this limitation, we present UAV-MM3D, a large-scale multi-modal, multi-scenario, and multi-weather synthetic dataset tailored for comprehensive model development, evaluation, and pretraining.

\begin{figure}
  \centering
  \includegraphics[width=0.48\textwidth]{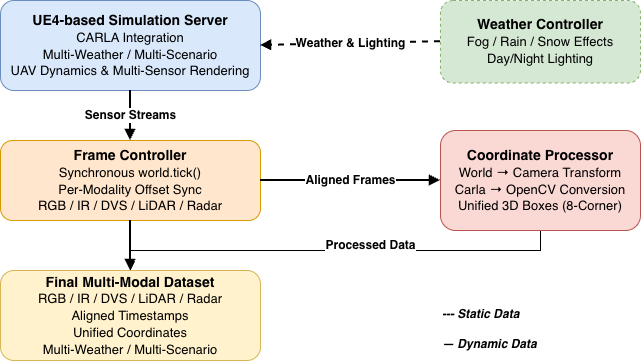}
  \caption{UAV-MM3D data collection framework. 
    UE4–CARLA simulation server provides multi-weather UAV scenes, 
    while the Python client consists of a Weather Controller, 
    Frame Controller, and Coordinate Processor. 
    It outputs a fully aligned multi-modal dataset across diverse conditions.}
  \label{sim_structure}
\end{figure}

\textbf{Collection Framework.}
Our multi-modal data collection framework adopts a modular architecture, as illustrated in Fig.~\ref{sim_structure}. It consists of two main components: (1) an Unreal Engine 4 (UE4)-based simulation server that integrates the CARLA~\cite{CARLA} platform, and (2) a Python client featuring two controllers and one processor—the Weather Controller, Frame Controller, and Coordinate Processor. The Weather Controller configures environmental parameters such as lighting and precipitation, the Frame Controller ensures temporal synchronization and data alignment across modalities, and the Coordinate Processor converts sensor outputs into a unified coordinate space for further processing.

\textbf{Sensor Settings.}
Our sensor configuration is designed to enable comprehensive ground-based observation of UAVs in diverse environmental conditions. The perception platform integrates multiple complementary modalities to achieve both geometric precision and visual robustness. We set up three different vision sensors: an RGB camera, an infrared (IR) camera, and a Dynamic Vision Sensor (DVS), each operating at a resolution of 1280 × 720 with a field of view of 90° (FoV). These vision sensors capture rich appearance and motion cues under varying illumination and dynamic flight patterns.

To enhance spatial awareness, 256-channel LiDAR and millimeter-wave radar are jointly placed. The LiDAR provides dense 3D measurements with a 120° horizontal and ± 20° vertical FoV, operating at 15 Hz and generating up to 2.5 million points per second. The radar, with a FoV of 120° × 40° × 40° and an effective range of 30 m, outputs the velocity, azimuth, and altitude for each detected object at 0.5 million points per second. Weather-dependent noise models are applied to both sensors to emulate realistic degradation in rain, fog, and snow.

\textbf{Scene Setting.}
We selected four representative simulated environments from the CARLA simulator—Town01, Town03HD, Town07HD and Town10HD—covering urban and suburban layouts with varying building densities. Data from each scene is from multiple ground viewpoints to ensure coverage diversity and cross-view generalization.

To evaluate robustness under environmental variations, data is collected under eight distinct weather conditions, including clear day/night, rain day/night, fog day/night, and snow day/night. Each weather setting adjusts physical parameters such as cloudiness, precipitation, fog density, and sun altitude angle within the CARLA~\cite{CARLA} physics-based atmosphere model, ensuring realistic visual and sensor-domain perturbations.

\begin{figure}
  \centering
  \includegraphics[width=0.48\textwidth]{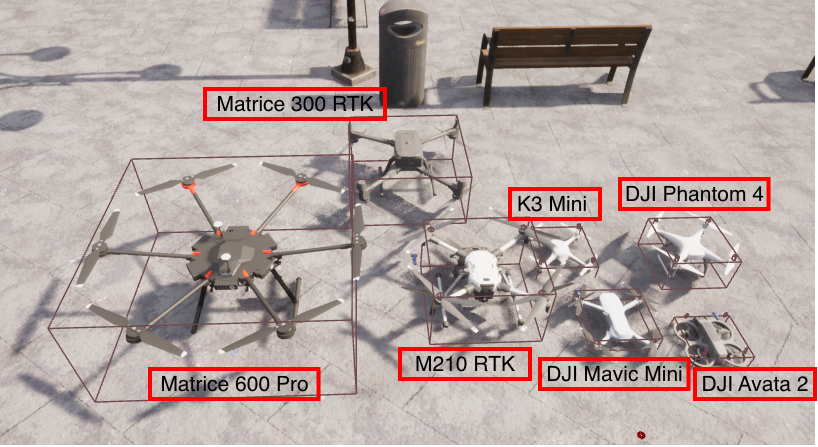}
    \caption{
    Visualization of the seven UAV platforms simulated in UAV-MM3D, including DJI Avata 2, DJI Mavic Mini, DJI Phantom 4, M210 RTK, Matrice 600 Pro, Matrice 300 RTK, and the K3 Mini drone. 
    }
  \label{show_drones}
\end{figure}

\textbf{Drone Setting.}
Within each scene–weather combination, we simulate 7 UAVs of different physical configurations, including DJI Avata 2, DJI Mavic Mini, DJI Phantom 4, M210 RTK, Matrice 600 Pro, Matrice 300 RTK, and K3 Mini Drone(Fig.~\ref{show_drones}). Each UAV is modeled according to its speed, mass and dimensions in the real-world, as shown in Table~\ref{tab:drone_specs}. In this way, we are able to make physically accurate flight dynamics such as acceleration, deceleration, turning, and roll tilting. This design enables the dataset to capture realistic flight postures and multi-UAV interactions across complex weather and illumination conditions.
\begin{table}[t]
\centering
\caption{Physical configurations of the seven UAV platforms in UAV-MM3D. Lightweight drones with incomplete public documentation use baseline parameters recorded in the dataset metadata.}
\setlength{\tabcolsep}{0.1pt}
\renewcommand{\arraystretch}{1.0}
\resizebox{\linewidth}{!}{
\begin{tabular}{lccc}
\toprule
\textbf{UAV Model} & \textbf{Mass} & \textbf{Dimensions (mm)} & \textbf{Max Speed} \\
\midrule
DJI Avata 2 & 0.377 kg & 185×212×64 & 27 m/s \\
DJI Mavic Mini & 0.249 kg & 159×202×55 & 13 m/s \\
DJI Phantom 4 & 1.38 kg & 289×289×196 & 20 m/s \\
Matrice 210 RTK V2 & 4.69 kg & 883×886×398 & 15 m/s \\
Matrice 600 Pro & 10.5 kg & 1668×1518×727 & 18 m/s \\
Matrice 300 RTK & 3.6 kg & 810×670×430 & 23 m/s \\
K3 Mini (E99) & 0.12 kg & 130×70×55 (folded) & 5 m/s \\
\bottomrule
\end{tabular}
}
\label{tab:drone_specs}
\end{table}

\textbf{Trajectory Generation.}
To simulate realistic UAV motion within each scene, we design a procedural waypoint generator implemented in Unreal Engine. For each scenario, up to seven UAVs are initialized with distinct physical and kinematic properties, and a unique set of waypoints is generated using an elliptical trajectory model. The trajectory is defined along the forward direction of a scene anchor (denoted as \textit{Locator}), forming a flattened oval path that gradually moves away from the ground sensors. Each trajectory consists of 15 to 25 discrete waypoints, evenly distributed along the parametric curve:
\begin{equation}
x = a \cos(\theta), \quad 
y = b \sin(\theta), \quad 
z = h_0 + \Delta h(\theta),
\end{equation}
where $a$ and $b$ control the longitudinal and lateral span of the ellipse, and $\Delta h(\theta)$ introduces stochastic height variation sampled within a predefined bound. The altitude and velocity profiles are randomized across UAVs to mimic diverse acceleration, deceleration, and turning behaviors.

\subsection{Data Processing}
All collected data are processed through a unified multi-threaded pipeline to ensure spatial and temporal consistency across all modalities. The pipeline performs synchronization and coordinate transformation.

\textbf{Spatio-temporal Alignment.}
Images, LiDAR point clouds, radar data, and DVS event streams are aligned using their simulation timestamps. Before each iteration of data capture, a world-tick function is invoked to synchronize all sensors within the same simulation frame, ensuring that every modality is updated under an identical timestamp. Frames are retained only when all modalities share a common timestamp, achieving pixel-level temporal consistency across sensors. 

\textbf{Coordinate Transformation.}
As summarized in Table~\ref{tab:coord_systems}, UAV-MM3D adopts six distinct coordinate frames to ensure consistent spatial representation across simulators and sensors. The CARLA world coordinate follows a left-handed ESU convention, whereas all physical sensors and image frames are defined in right-handed systems (e.g., FLU for LiDAR, RDF for cameras). All transformations between simulator, sensor, and image coordinates are stored in each sequence to maintain reproducible geometry and pixel-level alignment. After all, annotated 3D bounding boxes and projected multi-modal data are expressed in the camera-centric OpenCV coordinate system (RDF, right–down–forward), which serves as the unified reference frame for training, visualization, and evaluation.

\begin{table}[t]
\centering
\caption{Different Coordinate Systems in \textbf{UAV-MM3D}.}
\label{tab:coord_systems}
\small
\renewcommand{\arraystretch}{1.1}
\setlength{\tabcolsep}{3pt}
\resizebox{\linewidth}{!}{
\begin{tabular}{l|l|l|l}
\hline
\textbf{Name} & \textbf{Category} & \textbf{Type} & \textbf{Origin} \\
\hline
World & Global Ref. & ESU (L) & Fixed reference point \\
Ego & UAV Model & FLU (R) & Drone geometric center \\
Sensor--Cam & Image Sensor & RDF (R) & Camera optical center \\
Sensor--LiDAR & Range Sensor & FLU (R) & LiDAR center \\
Sensor--Radar & Motion Sensor & FRD (R) & Radar center \\
Simulator & CARLA Engine & ESU (L) & World origin \\
Rendered Img & OpenCV Frame & RDF (R, 2D) & Top-left of image \\
\hline
\end{tabular}}
\vspace{2pt}
\footnotesize{

\textbf{Note:} ESU: East–South–Up, FLU: Forward–Left–Up, FRD: Forward–Right–Down, RDF: Right–Down–Forward.
Handedness: R (Right-handed), L (Left-handed).
}
\end{table}

\begin{figure}
  \centering
  \includegraphics[width=0.48\textwidth]{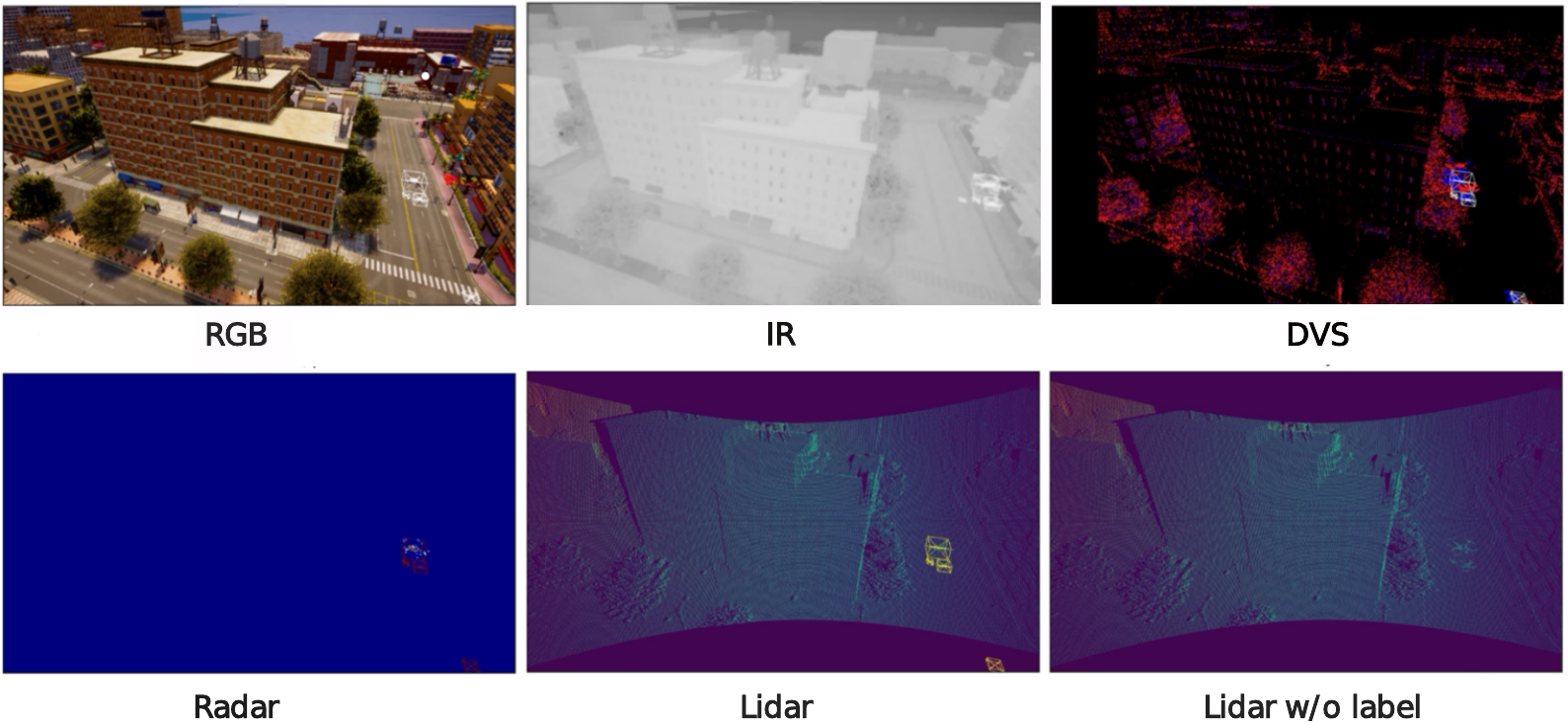}
    \caption{
    Examples of multi-modal sensor outputs in UAV-MM3D. 
    Each synchronized frame includes RGB imagery, infrared thermal data, and DVS event streams. 
    LiDAR point clouds are reprojected onto the RGB view for 2D alignment, while millimeter-wave radar measurements are rendered as velocity heatmaps. 
    All modalities are temporally aligned under a unified simulation timestamp.
    }
  \label{show_modalities}
\end{figure}

\subsection{Statistical Analysis}
In total, UAV-MM3D comprises approximately 1.1K simulated clips, corresponding to about 341K frames captured at a frame rate of 15 Hz.
Each clip spans roughly 20 seconds of continuous synchronized recording across all modalities, including RGB, infrared, DVS, LiDAR, and millimeter-wave radar, as illustrated in Fig.~\ref{show_modalities}.
Together, these clips yield over 1.7 million sensor images with 3D point clouds, forming a large-scale dataset for perception research under complex aerial-ground scenarios.

\begin{figure}[t]
  \centering
  \includegraphics[width=\linewidth]{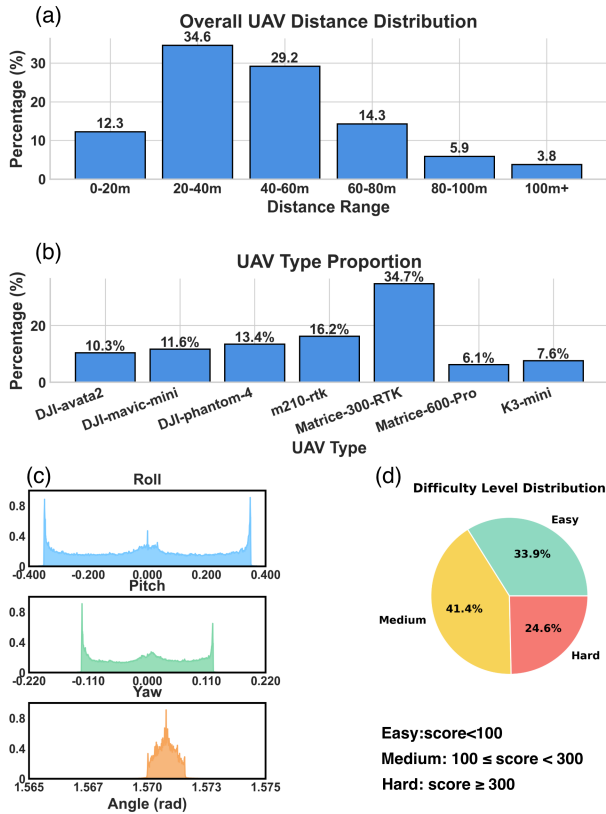}
  \caption{
    Distribution statistics of UAV-MM3D. 
    (a) Overall UAV–sensor distance distribution. 
    (b) Proportion of UAV types across all simulated scenes.
    (c) Orientation statistics (roll, pitch, yaw) computed from 3D bounding boxes.
    (d) Difficulty distribution estimated from UAV size, sensing distance, and weather conditions.
  }
  \label{drone_distribution}
\end{figure}


Each simulated scene involves \textbf{one to seven UAVs} with heterogeneous physical models and flight dynamics, producing substantial variation in scale, appearance, and occlusion across viewpoints.
As shown in Fig.~\ref{drone_distribution}(a), UAV types are well balanced across the dataset: the Matrice-300-RTK appears most frequently due to its representative size and dynamics, while extremely large or small platforms naturally occur less often.
Fig.~\ref{drone_distribution}(b) further reports the distribution of UAV–sensor distances, where most cases fall within 20–60 meters, accompanied by both close-range and long-range examples that together form a realistic and diverse depth profile.
Finally, Fig.~\ref{drone_distribution}(c) shows the orientation statistics, which exhibit broad roll and pitch variation but a narrow yaw range. This reflects realistic multirotor flight dynamics—real UAVs primarily rely on roll-based banking rather than large heading changes during lateral motion.

\textbf{Data Categorization.}  
To enable comprehensive evaluation under varying perception challenges, all annotated samples in UAV-MM3D are divided into three difficulty levels: \textit{easy}, \textit{moderate}, and \textit{hard}.  
The difficulty score is defined as the ratio between the UAV’s distance and its physical size, scaled by a weather-dependent weighting factor:
 \begin{equation}
    S = \frac{d}{s} \times w_{\text{weather}},
 \end{equation}
where \(d\) denotes the camera-to-UAV distance, \(s\) represents the UAV’s maximum physical dimension, and \(w_{\text{weather}}\) corresponds to the visibility degradation (e.g., 1.0 for clear, 2.5 for fog).  
This stratified organization ensures balanced coverage across different visibility and scale conditions, facilitating robust model benchmarking under diverse sensing scenarios, as illustrated in Fig.~\ref{drone_distribution}(d).

\begin{figure}
  \centering
  \includegraphics[width=0.48\textwidth]{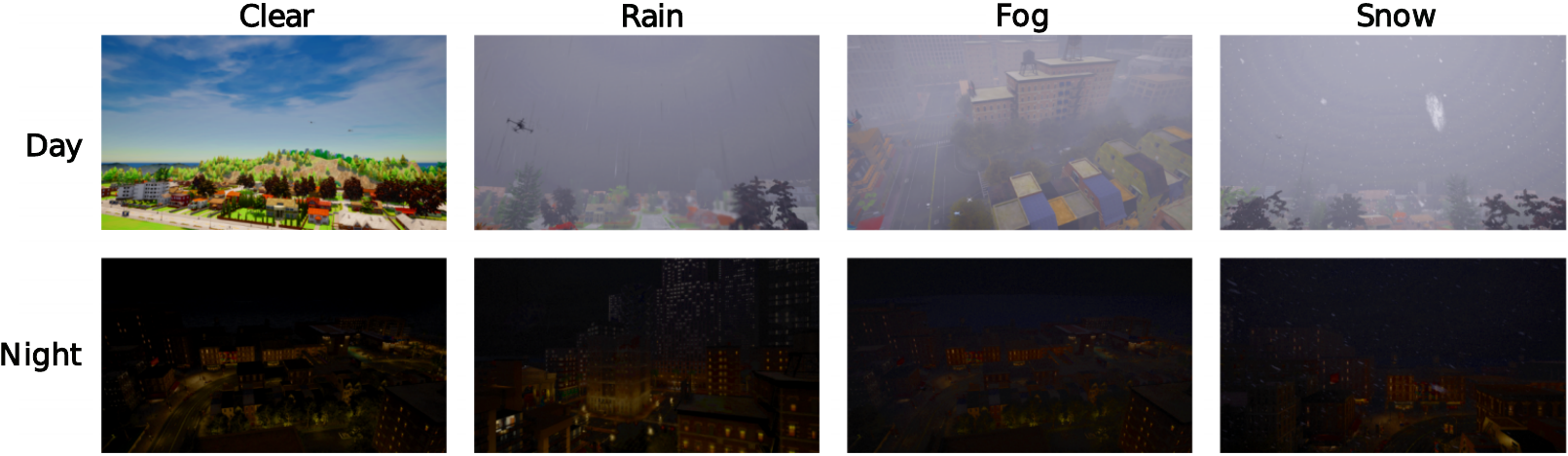}
    \caption{
    Illustration of the eight physically based weather conditions in UAV-MM3D. 
    Each environment is rendered under both day and night illumination, covering clear, rain, fog, and snow settings. 
    These combinations introduce diverse lighting, atmospheric, and visibility variations that enhance the realism and challenge of UAV perception across different urban and suburban scenes.
    }
  \label{show_weathers}
\end{figure}

\textbf{Scene Statistics.}
The UAV-MM3D dataset encompasses a broad spectrum of simulation conditions, integrating four representative CARLA towns with distinctive urban and suburban topologies and eight physically based weather configurations that modulate illumination, precipitation, and atmospheric visibility, as illustrated in Fig.~\ref{show_weathers}.
These environments jointly provide diverse visual and geometric contexts that reflect varying levels of structural complexity, traffic density, and environmental dynamics.

\section{Applications of UAV-MM3D}
UAV-MM3D provides a unified and comprehensive evaluation environment for a wide range of Anti-UAV perception tasks, including 2D detection, 3D detection, 6-DoF pose estimation, multi-object tracking, and short-term trajectory prediction. Among these tasks, 6-DoF pose estimation plays a central role because accurate UAV localization and orientation are essential prerequisites for downstream tracking and interception strategies. In this section, we present the multimodal 6-DoF pose estimation task supported by UAV-MM3D, followed by the proposed LiDAR-guided baseline LGFusionNet. We further introduce three multimodal 3D detection baselines and a series of ablation studies.

\subsection{UAVs Multimodal 6-DoF Pose Estimation}
\subsubsection{Task Definition}
The objective of multimodal 6-DoF pose estimation is to recover the full 3D position $(x,y,z)$ and 3D rotation $(\theta, \phi, \psi)$ of UAVs from synchronized multimodal sensor streams. Compared to ground vehicles or large-scale objects, UAVs in low-altitude airspace exhibit several unique challenges: (1) extremely fast motion and abrupt trajectory changes; (2) small physical size and low pixel footprint at medium-to-long distances; (3) severe viewpoint variation due to free-space 3D maneuvering; and (4) strong modality inconsistency across RGB, IR, Radar, and LiDAR caused by illumination changes or sparse geometry. UAV-MM3D offers temporally aligned multimodal data, allowing systematic exploration of cross-modal feature correspondence and uncertainty reduction in pose estimation.

\subsubsection{LGFusionNet Architecture}
\begin{figure*}[t]
    \centering
    \includegraphics[width=1\linewidth]{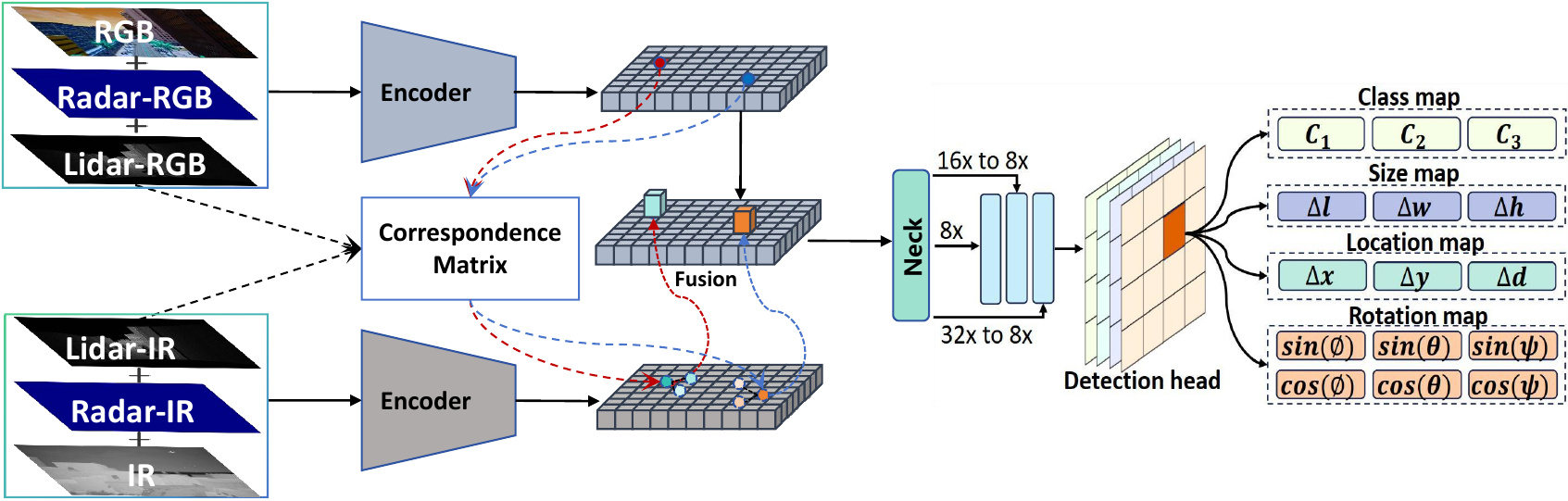}
    \caption{
    Overall architecture of the proposed LGFusionNet. The network adopts a dual-branch multi-sensor feature encoding design, where the RGB and IR branches independently encode their stacked inputs (image, LiDAR-projected depth, and Radar-projected velocity) using two separate ResNet-50 backbones with identical architecture.  
    LiDAR points are projected onto both RGB and IR views to construct cross-view geometric correspondences, which are then used by the LiDAR-guided cross-branch spatial alignment module. This module samples IR-branch features according to the RGB–IR projection pairs and aggregates them via KNN-based neighborhood fusion to obtain aligned IR descriptors.      
    The aligned IR-branch features are then fused with RGB-branch features in the RGB coordinate space through channel-wise concatenation followed by convolution, producing a spatially consistent multimodal representation. This fused representation goes through a neck to the prediction head, supporting both 6-DoF UAV pose estimation and object classification as downstream tasks. Specifically, rotations are predicted using Direction-Angle Parameterization (DAP) for stable and continuous angular regression.
    }

    \label{fig:lgfusionnet}
\end{figure*}

To exploit the complementary strengths of different modalities, we introduce \textbf{LGFusionNet} — a LiDAR-guided multimodal 6-DoF pose estimation network (architecture shown in Fig.~\ref{fig:lgfusionnet}).

\paragraph{Two-Branch Multi-Sensor Feature Encoding.}
LGFusionNet adopts a dual-branch structure to handle heterogeneous modalities:

\begin{itemize}
\item \textbf{RGB Branch}: Processes appearance-rich inputs (RGB imagery + LiDAR-projected depth + Radar-projected velocity) via a ResNet-50 backbone. LiDAR/Radar projections are mapped to the RGB image plane using RGB camera extrinsic parameters, providing geometric/motion priors for texture-poor scenarios. The branch feature is
  \begin{equation}
      F_{\text{RGB-branch}} = \text{Concat}\left( F_{\text{RGB}}, F_{\text{LiDAR(RGB)}}, F_{\text{Radar(RGB)}} \right),
  \end{equation}
  where $F_{\text{RGB}}$, $F_{\text{LiDAR(RGB)}}$, $F_{\text{Radar(RGB)}}$ denote features of RGB imagery, LiDAR-projected depth, and Radar-projected velocity (RGB branch), respectively.

\item \textbf{IR Branch}: Processes illumination-invariant inputs (IR imagery + LiDAR-projected depth + Radar-projected velocity) via the same ResNet-50 backbone. Projections use IR camera extrinsic parameters, and the branch feature is
  \begin{equation}
      F_{\text{IR-branch}} = \text{Concat}\left( F_{\text{IR}}, F_{\text{LiDAR(IR)}}, F_{\text{Radar(IR)}} \right),
  \end{equation}
  where $F_{\text{IR}}$, $F_{\text{LiDAR(IR)}}$, $F_{\text{Radar(IR)}}$ correspond to IR branch features.
\end{itemize}

\paragraph{LiDAR-Guided Cross-Branch Spatial Alignment.}
To establish geometric consistency between the two multimodal branches rather than raw RGB/IR images, LGFusionNet performs LiDAR-guided alignment between $F_{\text{RGB-branch}}$ and $F_{\text{IR-branch}}$. Since each branch internally incorporates image (RGB or IR), LiDAR-projected depth, and Radar-projected velocity cues, LiDAR points serve as a stable geometric anchor to bridge the heterogeneous representation spaces of the two branches.

\begin{itemize}
\item \textbf{LiDAR-Based Cross-View Point Correspondence Construction.}  
LiDAR points are defined in world coordinates and can be accurately projected to both RGB and IR cameras using their intrinsic and extrinsic parameters. For a LiDAR point $P_k = (X_k, Y_k, Z_k)^\top$, the projection onto camera $c \in \{\text{RGB}, \text{IR}\}$ is
\begin{equation}
    p_{c,k} \sim K_c [R_c \,|\, t_c],
\end{equation}
where $p_{c,k} = (u_{c,k}, v_{c,k})^\top$ denotes the pixel coordinates of $P_k$ in camera $c$'s image plane, $K_c$ is the intrinsic matrix of camera $c$, and $[R_c|t_c]$ represents its extrinsic parameters that transform 3D points from world to camera coordinates. By projecting the same LiDAR point to RGB and IR cameras, we obtain correspondence pairs $(\pi_{\text{RGB}}(P_k), \pi_{\text{IR}}(P_k))$, providing sparse but geometrically reliable anchors across the RGB-branch and IR-branch feature spaces.

\item \textbf{Projection-Guided IR-Branch Feature Sampling and KNN Aggregation.}  
Let $F_{\text{RGB-branch}}, F_{\text{IR-branch}} \in \mathbb{R}^{H \times W \times D}$ denote the two branch feature maps at the same resolution. For each pixel $(i,j)$ in $F_{\text{RGB-branch}}$, we collect the corresponding IR projections whose RGB projections fall in the local neighborhood $\mathcal{N}((i,j); r)$:
\[
S_{i,j} = \left\{ \pi_{\text{IR}}(P_k) \,\big|\, \pi_{\text{RGB}}(P_k)\in\mathcal{N}((i,j); r) \right\}.
\]
Features of the IR-branch are retrieved at the projected positions and aggregated using $K$-nearest neighbors:

\begin{equation}
    \mathcal{I}_{i,j} \leftarrow \text{KNN}(S_{i,j}, K),
\end{equation}
\begin{equation}
    \hat{f}_{\text{IR-branch}}(i,j) =
    \begin{cases}
        \mathcal{A}\big(F_{\text{IR-branch}}, \mathcal{I}_{i,j}\big), & S_{i,j} \neq \emptyset, \\[2pt]
        \mathbf{0}, & \text{otherwise},
    \end{cases}
\end{equation}

where $\hat{f}_{\text{IR-branch}}(i,j) \in \mathbb{R}^D$ denotes the aligned IR feature at pixel $(i,j)$, 
and $\mathcal{I}_{i,j}$ is the set of $K$ nearest neighbors in the IR-branch feature map corresponding to LiDAR points projected within the neighborhood of RGB-branch pixel $(i,j)$.  
Here, $\mathcal{A}(\cdot, \cdot)$ represents the aggregation operator (e.g., averaging) over the selected features.

\item \textbf{Cross-Branch Feature Fusion on RGB-Branch Coordinate Space.}  
The aligned IR-branch feature is fused with the RGB-branch feature at each pixel by channel-wise concatenation followed by a convolution:
\begin{equation}
    f_{\text{cat}}(i,j) =
    \text{Concat}\!\big(f_{\text{RGB-branch}}(i,j),\, \hat{f}_{\text{IR-branch}}(i,j)\big),
\end{equation}
\begin{equation}
    f_{\text{fused}}(i,j) =
    \mathrm{Conv}\!\big( f_{\text{cat}}(i,j) \big),
\end{equation}
where $f_{\text{cat}}(i,j)$ denotes the concatenated RGB–IR feature vector at pixel $(i,j)$ prior to convolutional fusion.
This convolutional operation produces the cross-branch fused feature map 
$F_{\text{fused}} \in \mathbb{R}^{H \times W \times D'}$, 
which enhances RGB-branch perception with illumination-robust IR cues while preserving geometric correctness guaranteed by LiDAR projections. 
The overall LiDAR-guided cross-branch alignment and feature fusion pipeline is summarized in Algorithm~\ref{alg:lidar_fusion}.
\end{itemize}

\begin{algorithm}[t]
\caption{LiDAR-Guided RGB--IR Branch Alignment and Feature Fusion}
\label{alg:lidar_fusion}
\begin{algorithmic}[1]
\Require \\
LiDAR points $\{P_k\}$; \\ 
intrinsics/extrinsics $K_c,[R_c|t_c]$; \\
feature maps $F_{\text{RGB-branch}}, F_{\text{IR-branch}}$; \\
radius $r$; \\
neighbors $K$
\Ensure Fused feature map $F_{\text{fused}}$

\For{each LiDAR point $P_k$}
    \State $p_{\text{RGB}} \leftarrow \text{Project}(P_k, K_{\text{RGB}}, [R_{\text{RGB}}|t_{\text{RGB}}])$
    \State $p_{\text{IR}} \leftarrow \text{Project}(P_k, K_{\text{IR}}, [R_{\text{IR}}|t_{\text{IR}}])$
    \State store correspondence pair $(p_{\text{RGB}}, p_{\text{IR}})$
\EndFor

\For{each pixel $(i,j)$ in $F_{\text{RGB-branch}}$}
    \State $S_{i,j} \leftarrow \{ p_{\text{IR}} \mid p_{\text{RGB}}\in\mathcal{N}((i,j); r) \}$
    \If{$S_{i,j} = \emptyset$}
        \State $\hat{f}_{\text{IR-branch}}(i,j) \leftarrow \mathbf{0}$
    \Else
        \State $\mathcal{I}_{i,j} \leftarrow \text{KNN}(S_{i,j}, K)$
        \State $\hat{f}_{\text{IR-branch}}(i,j) \leftarrow \text{Aggregate}(F_{\text{IR-branch}}, \mathcal{I}_{i,j})$

    \EndIf
    \State $f_{\text{cat}}(i,j) \leftarrow \text{Concat}(f_{\text{RGB-branch}}(i,j),\, \hat{f}_{\text{IR-branch}}(i,j))$
    \State $f_{\text{fused}}(i,j) \leftarrow \mathrm{Conv}(f_{\text{cat}}(i,j))$

\EndFor

\State \Return $F_{\text{fused}}$
\end{algorithmic}
\end{algorithm}

\paragraph{Remarks.}
This LiDAR-guided procedure ensures that cross-branch fusion is performed between $F_{\text{RGB-branch}}$ and $F_{\text{IR-branch}}$ at geometrically corresponding spatial locations, rather than relying on appearance similarity alone. Such spatially enforced alignment is crucial given the viewpoint differences between the RGB and IR cameras and the heterogeneous sensing characteristics of the associated LiDAR/Radar projections within each branch. In addition, the KNN-based aggregation on the IR feature map improves robustness to the sparsity of LiDAR projections and alleviates discretization artifacts near depth discontinuities, ensuring stable multimodal fusion in long-range or low-visibility scenarios.

\subsubsection{Comparative Baselines}
To comprehensively assess the contribution of LiDAR-guided cross-modal alignment in LGFusionNet, we design comparative baselines based on two representative technical paradigms. Each paradigm is evaluated under three modal configurations (RGB-only, IR-only, Score-Level Fusion) to isolate the impact of fusion strategies, with all baselines implemented within the same unified framework:

\begin{itemize}
    \item \textbf{YOLO-6D} \cite{YOLO6D}:
    A geometric-based paradigm that inherits the core idea of Keypoint+PnP—extracting 2D keypoints from input data, followed by the PnP (Perspective-n-Point) algorithm for pose estimation. It provides a lower bound for geometry-driven pose inference, with three modal settings:
    
    - RGB-only: Uses only RGB inputs (texture-rich but illumination-sensitive);
    
    - IR-only: Uses only IR inputs (illumination-robust but lacking fine-grained appearance cues);
    
    - Score-Level Fusion (Late Fusion): Decision-level fusion of independent detectors trained on four unaligned modalities—RGB, IR, LiDAR projection, and Radar projection—via weighted averaging. Modalities are not spatially aligned prior to fusion, relying solely on post-hoc score merging.

    \item \textbf{Center-based Regression} \cite{CenterNet}:
    A modern one-stage paradigm that models pose estimation as center-point heatmap regression, representing state-of-the-art monocular pose approaches. It also adopts three modal configurations:
    
    - RGB-only: RGB inputs as the sole input;
    
    - IR-only: IR inputs as the sole input;
    
    - Score-Level Fusion (Late Fusion): Decision-level weighted fusion of regression results from four unaligned modalities (RGB, IR, LiDAR projection, Radar projection), with no prior spatial alignment between heterogeneous sensors.

For all Late Fusion variants in the baselines, weighted averaging is applied to the final detection confidence scores from each modality branch:
\begin{equation}
    S_{\text{final}} = \alpha_1 S_{\text{RGB}} + \alpha_2 S_{\text{IR}} + \alpha_3 S_{\text{LiDAR\text{-}Proj}} + \alpha_4 S_{\text{Radar\text{-}Proj}},
\end{equation}
where $S_{\text{RGB}}$, $S_{\text{IR}}$, $S_{\text{LiDAR\text{-}Proj}}$, and $S_{\text{Radar\text{-}Proj}}$ denote the modality-specific confidence scores for object existence/classification. The coefficients satisfy $\alpha_1 + \alpha_2 + \alpha_3 + \alpha_4 = 1$, and are fixed as $\alpha_1 = 0.3$, $\alpha_2 = 0.3$, $\alpha_3 = 0.2$, $\alpha_4 = 0.2$. The fused score \(S_{\text{final}}\) is used for detection ranking and NMS, while regression outputs (center, rotation, and size) are taken from the modality branch with the highest confidence to avoid instability caused by averaging heterogeneous regression predictions.

    For comparison, LGFusionNet's own Late Fusion variant adopts a distinct design: it fuses results from two dedicated branches (RGB-branch and IR-branch) rather than directly merging four independent modalities. Each branch inherently integrates its core modality (RGB/IR) with LiDAR projection and Radar projection as auxiliary geometric cues. This variant serves as an additional ablation to verify whether LiDAR-guided spatial alignment (the core innovation) provides incremental value beyond the structured branch design.
\end{itemize}

\subsubsection{Pose Decoding and Metrics}
All methods predict full 6-DoF pose parameters—including 3D translation $(x,y,z)$, rotation $(\theta,\phi,\psi)$, size $(w,h,l)$—together with 2D and 3D bounding boxes. Performance is evaluated using standard metrics:

\begin{itemize}
    \item Rotation error (degrees), Position error (meters), Size error (meters);
    \item 2D Average Precision (2D AP), 3D Average Precision (3D AP).
\end{itemize}

\subsubsection{Results}
Although the UAV-MM3D dataset includes a DVS (event camera) modality, it is excluded from our main fusion pipeline. In rain/fog and nighttime scenes, illumination flickering and reflections often trigger dense pseudo-events, yielding unstable temporal statistics and weak geometric correspondence with frame-based modalities. Preliminary experiments showed that even after event denoising and temporal resampling, incorporating DVS degraded pose supervision and overall accuracy. To ensure fair comparison and focus on the effect of LiDAR-guided fusion, LGFusionNet is evaluated using RGB–IR–LiDAR–Radar inputs. Nonetheless, the high dynamic range and motion sensitivity of DVS remain valuable, and developing robust event–frame alignment under adverse weather will be explored in future work.

\textbf{Implementation details.}
All detectors were trained on 8$\times$RTX 4090 GPUs using the Adam optimizer~\cite{Adam} with an initial learning rate of 0.001 and cosine decay. LGFusionNet was trained for 50 epochs on UAV-MM3D with a batch size of 4 per GPU. For the LiDAR-guided alignment module, we set the projection neighborhood radius to $r=4$ pixels and used $K=5$ nearest neighbors for local feature aggregation. Both RGB and IR branches employ ResNet-50 backbones with identical initialization and synchronized multimodal augmentations (random horizontal flip, brightness jitter, and depth/velocity dropout).

\begin{table*}[t]
\centering
\caption{Comparison of multimodal and unimodal 6-DoF pose estimation methods on the UAV-MM3D dataset. 
LGFusionNet achieves the best overall performance by leveraging LiDAR-guided cross-modal alignment. 
Center\&DAP Encoding denotes the combination of Center Encoding (for target center localization) and DAP Encoding (for rotation angle parameterization), addressing spatial positioning and angular regression in 6-DoF pose estimation respectively.
Note: $\dagger$ indicates Score-Level Fusion of four unaligned modalities (RGB, IR, LiDAR projection, Radar projection); 
$\ddagger$ indicates Score-Level Fusion of two branches (RGB-branch and IR-branch), each integrating RGB/IR with LiDAR/Radar projections.}
\label{tab:pose_estimation_baselines}
\renewcommand{\arraystretch}{1.3} 
\resizebox{\linewidth}{!}{%
\begin{tabular}{lcccccccc}
\toprule
\textbf{Method} & \textbf{Modality Fusion} & \textbf{Target Encoding} & \textbf{Backbone} &
\textbf{Rot. Err$\downarrow$} & \textbf{Pos. Err$\downarrow$} & \textbf{Size Err$\downarrow$} &
\textbf{2D AP$\uparrow$} & \textbf{3D AP$\uparrow$} \\
\midrule
\multirow{3}{*}{YOLO-6D \cite{YOLO6D}} & RGB-only & Keypoint Encoding & ResNet50 &
26.47 & 7.21 & 0.43 & 36.44 & 11.78 \\
& IR-only & Keypoint Encoding & ResNet50 &
24.15 & 6.83 & 0.41 & 38.92 & 12.26 \\
& Late Fusion$\dagger$ & Keypoint Encoding & ResNet50 &
22.73 & 5.95 & 0.40 & 40.51 & 13.89 \\
\midrule
\multirow{3}{*}{\parbox{2.3cm}{Center-based Regression \cite{CenterNet}}} & RGB-only & Center Encoding & ResNet50 &
21.65 & 6.48 & 0.42 & 41.28 & 13.07 \\
& IR-only & Center Encoding & ResNet50 &
19.29 & 5.76 & 0.40 & 42.15 & 14.64 \\
& Late Fusion$\dagger$ & Center Encoding & ResNet50 &
17.84 & 4.92 & 0.39 & 43.67 & 17.38 \\
\midrule
\multirow{2}{*}{\parbox{2cm}{\centering LGFusionNet (Ours)}}
& Late Fusion$\ddagger$ & Center\&DAP Encoding & ResNet50 &
12.48 & 3.95 & 0.37 & 43.89 & 23.11 \\
& LiDAR-guided Fusion & Center\&DAP Encoding & ResNet50 &
\textbf{10.57} & \textbf{3.64} & \textbf{0.35} &
\textbf{45.35} & \textbf{26.72} \\
\bottomrule
\end{tabular}%
}
\end{table*}

\textbf{Evaluation results.} Table~\ref{tab:pose_estimation_baselines} summarizes the comparative results across different technical paradigms and modal configurations. 
Across both YOLO-6D \cite{YOLO6D} and Center-based Regression paradigms, a consistent trend emerges: IR-only unimodal models outperform their RGB-only counterparts, thanks to IR's inherent robustness to low-illumination conditions in low-altitude UAV scenes. Score-Level Fusion (Late Fusion) further improves upon single-modality performance for all baselines, as it leverages complementary appearance and thermal cues. However, Late Fusion's gains remain limited—its lack of spatial alignment between RGB and IR modalities leads to suboptimal feature interaction, especially under fast UAV motion and viewpoint variations.  

Among the baseline paradigms, Center-based Regression \cite{CenterNet} consistently outperforms YOLO-6D \cite{YOLO6D} across all modal configurations. This is attributed to its end-to-end regression design, which better captures fine-grained pose correlations compared to YOLO-6D's two-stage geometric inference (prone to keypoint detection errors, inheriting the limitation of the Keypoint+PnP approach). Even so, the best baseline (Center-based Regression \cite{CenterNet} with Late Fusion) still struggles to mitigate cross-modal misalignment, resulting in relatively high rotation error (17.84\textdegree) and position error (4.92 m).  

In stark contrast, LGFusionNet demonstrates superior performance across all metrics, with its core LiDAR-guided Fusion configuration achieving the best results. Compared to the top-performing baseline, LGFusionNet reduces position error by more than \textbf{25\%} (from 4.92 m to 3.64 m) and rotation error by over \textbf{40\%} (from 17.84\textdegree to 10.57\textdegree), while boosting 3D AP by more than \textbf{50\%} (from 17.38 to 26.72). Notably, LGFusionNet's own Late Fusion variant also outperforms all baseline methods, confirming the effectiveness of its dual-branch encoding architecture. The additional gains from LiDAR-guided spatial alignment further validate that geometry-driven cross-modal correspondence is critical for resolving ambiguities in complex low-altitude scenes—this is the key advantage that distinguishes LGFusionNet from traditional fusion approaches.

\subsubsection{Ablation Studies}
We further analyze the impact of key components in our framework.

\textbf{Effect of Input Resolution.}
Since the perceived object scale in monocular 3D detection is strongly influenced by image resolution, we replace the original focal-length ablation with a resolution-based study. We evaluate three commonly used input resolutions $\{960 \times 540, 1060 \times 600, 1280 \times 720\}$ while keeping all camera intrinsics unchanged. As shown in Table~\ref{tab:resolution_ablation}, input resolution has a \textbf{significant impact} on pose estimation performance: with the gradual increase of resolution, all evaluation metrics are continuously optimized. Specifically, compared with the lowest resolution (960×540), the optimal resolution (1280×720) reduces rotation error by 3.85° (14.42°→10.57°), position error by 0.84 m (4.48 m→3.64 m), and size error by 0.06 (0.41→0.35), while boosting 3D AP by 6.57 (20.15→26.72) — a relative improvement of over \textbf{32\%}. This is attributed to higher resolution providing clearer object boundaries, richer texture details, and more precise geometric cues, which are critical for accurate 6D pose inference. The consistent performance gain across resolution scales confirms that our model can effectively leverage the additional information from high-resolution inputs, and the 1280×720 configuration achieves the best balance between precision and computational efficiency.

\begin{table}[t]
\centering
\caption{Ablation study on input image resolution. 
Higher resolution provides richer geometric and texture cues, leading to continuous performance improvement.}
\label{tab:resolution_ablation}
\resizebox{\linewidth}{!}{
\begin{tabular}{ccccc}
\toprule
\textbf{Resolution} & \textbf{Rot. Error$\downarrow$} &
\textbf{Pos. Error$\downarrow$} &
\textbf{Size Error$\downarrow$} &
\textbf{3D AP$\uparrow$} \\
\midrule
960$\times$540 & 14.42 & 4.48 & 0.41 & 20.15 \\
1060$\times$600 & 12.98 & 3.92 & 0.39 & 23.32 \\
1280$\times$720 & \textbf{10.57} & \textbf{3.64} & \textbf{0.35} & \textbf{26.72} \\
\bottomrule
\end{tabular}
}
\end{table}



\textbf{Effect of LiDAR-Guided Cross-Branch Fusion.}
To investigate the contribution of each component in LGFusionNet, we conduct ablation experiments by progressively enhancing the fusion strategy and introducing the LiDAR-guided correspondence module. As shown in Table~\ref{tab:lgfusion_ablation}, the RGB-branch and IR-branch variants exhibit complementary strengths: the RGB branch achieves better spatial precision under normal lighting, while the IR branch is more robust in low-illumination or high-glare conditions.

We further explore two fusion paradigms: \textit{Late Fusion} (score-level merging of independent branch outputs) and \textit{Mid-Attention Fusion} (feature-level interaction via cross-attention before decoding). The results show that Mid-Attention Fusion outperforms Late Fusion, as feature-level interaction captures finer-grained cross-modal correlations compared to post-hoc score averaging. However, both fusion strategies without LiDAR guidance still suffer from heterogeneous modality misalignment, leading to suboptimal performance.

In contrast, the full model with LiDAR-guided correspondence achieves the best performance across all metrics. This demonstrates that LiDAR serves as a reliable geometric anchor to align cross-sensor features, and when combined with mid-level attention fusion, it maximizes the complementary value of RGB and IR modalities.

\begin{table}[t]
\centering
\caption{
Ablation of LGFusionNet components. 
RGB-branch = RGB branch only; IR-branch = IR branch only; 
Dual-LateFusion = dual branches with Late Fusion (score-level); 
Dual-MidAttn = dual branches with Mid-Attention Fusion (feature-level); 
Full (LGF) = full model with LiDAR-guided mid-attention fusion.
}
\label{tab:lgfusion_ablation}
\resizebox{\linewidth}{!}{
\begin{tabular}{lcccc}
\toprule
Model Variant & \textbf{Rot. Err$\downarrow$} & \textbf{Pos. Err$\downarrow$} & \textbf{Size Err$\downarrow$} & \textbf{3D AP$\uparrow$} \\
\midrule
RGB-branch    & 18.92 & 4.78 & 0.42 & 14.86 \\
IR-branch     & 15.64 & 4.22 & 0.41 & 18.93 \\
Dual-LateFusion     & 12.48 & 3.95 & 0.37 & 23.11 \\
Dual-MidAttn  & 11.75 & 3.82 & 0.37 & 24.68 \\
\rowcolor{gray!10}
Full (LGF)    & \textbf{10.57} & \textbf{3.64} & \textbf{0.35} & \textbf{26.72} \\
\bottomrule
\end{tabular}
}
\end{table}

\subsection{UAVs Target Tracking}
Reliable UAV target tracking is essential for accurate low-altitude perception and serves as a critical foundation for downstream tasks such as trajectory prediction and intent understanding. Our dataset offers large-scale, high-fidelity, and multi-modal observations, providing comprehensive support for advancing UAV tracking research under challenging conditions, including appearance variation, and scale changes.

\begin{table}[ht]
\centering
\begin{tabular}{lccc}
\hline
Method & AUC $\uparrow$ & $P$ $\uparrow$ & $P_{\mathrm{norm}}$ $\uparrow$ \\
\cline{2-3}
\hline
OSTrack~\cite{OSTrack} & 21.8 & 51.9 & 41.3 \\
DropTrack~\cite{DropTrack} & 28.8 & 53.7 & 41.3 \\
ARTrack~\cite{ARTrack} & 29.8 & 58.0 & 42.9 \\
ODTrack~\cite{ODTrack} & \bf{30.7} & \bf{58.1} & \bf{44.9}\\
\hline
\end{tabular}
\caption{UAV target tracking performance on UAV-MM3D dataset.}
\label{tab:UAVsTrack}
\end{table}

\begin{figure*}[ht]
  \centering
  \includegraphics[width=0.8\textwidth]{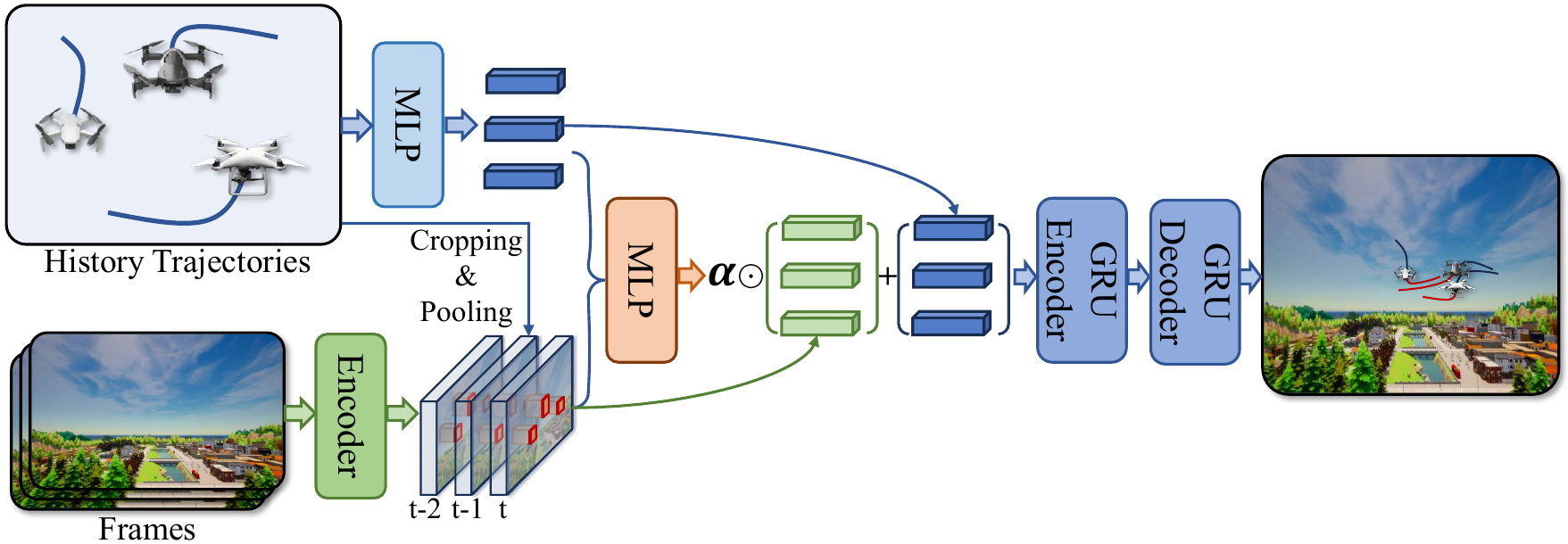}
  \caption{The framework of the proposed UAVsTrajectory. Our framework takes historical 3D UAV trajectories and corresponding RGB images as inputs. The historical trajectories and image frames are first encoded separately. Guided by the 2D bounding boxes, target regions are cropped from the images and pooled to obtain compact appearance representations. An adaptive fusion module then learns modality-wise weighting coefficients to integrate trajectory and appearance features. Finally, the fused representation is decoded to predict future UAV trajectories.}
  \label{UAVsTraj}
\end{figure*}

\begin{table*}[ht]
\centering
\renewcommand{\arraystretch}{1.2}
\resizebox{0.8\linewidth}{!}{ 
\begin{tabular}{lcccccc}
\hline
Method & ADE@1s $\downarrow$ & FDE@1s $\downarrow$ & ADE@3s $\downarrow$ & FDE@3s $\downarrow$& ADE@5s $\downarrow$ & FDE@5s $\downarrow$\\
\cline{2-5}
\hline
Kalman Filter~\cite{KF} & 5.73 & 9.85 & 17.10 & 33.50 & 29.27 & 58.96 \\
LSTM~\cite{lstm} & 2.45 & 4.97 & 8.86 & 17.78 & 14.93 & 27.84 \\
\bf{Ours} & \bf{2.37} & \bf{4.32} & \bf{8.58} & \bf{17.29} & \bf{14.25} & \bf{26.94}\\
\hline
\end{tabular}
}
\caption{Trajectory prediction performance (ADE / FDE in meters) on UAV-MM3D dataset. We evaluate the proposed method over prediction horizons of 1 s, 3 s, and 5 s. ADE measures the average per-frame displacement error over the prediction interval, while FDE denotes the displacement error at the final predicted timestep. We compare our approach with a traditional Kalman filter–based method and an LSTM-based learning model.}
\label{tab:UAVsTrajectory}
\end{table*}

\textbf{Metrics.} We report the following widely used metrics for the target tracking task. Area Under the Curve (AUC) measures the area under the success plot, which represents the fraction of frames whose Intersection-over-Union (IoU) exceeds varying thresholds as a function of the overlap threshold. Precision ($P$) evaluates the percentage of frames in which the predicted bounding-box center lies within a specified distance (typically 20 pixels) from the ground-truth center. Normalized Precision ($P_{\mathrm{norm}}$) normalizes the center error by the size of the ground-truth bounding box, providing a scale-invariant measurement. Higher values of AUC, $P$, and $P_{\mathrm{norm}}$ indicate better tracking performance. 

\textbf{Baselines.} We evaluate our dataset using four representative and widely adopted target tracking methods: OSTrack~\cite{OSTrack}, DropTrack~\cite{DropTrack}, ARTrack~\cite{ARTrack}, and ODTrack~\cite{ODTrack}. These methods cover diverse and strong tracking paradigms and are chosen to provide a comprehensive and fair evaluation of tracking performance on our dataset. All methods are implemented using their official configurations and trained under the same experimental protocol for fair comparison.

\textbf{Results.} Table~\ref{tab:UAVsTrack} reports the tracking performance of different baselines on our dataset. Existing methods achieve reasonable results in AUC and precision, but all exhibit noticeable performance degradation compared to standard benchmarks, indicating the increased difficulty of low-altitude UAV tracking. The results demonstrate that our dataset is challenging and well-suited for evaluating robust UAV target tracking methods.

Future work will explore multi-modal fusion of complementary sensing cues, including RGB, infrared, LiDAR, and radar data, to enable more accurate and robust low-altitude UAV target tracking.

\subsection{UAVs Trajectory Prediction}
Accurate prediction of low-altitude UAVs trajectories is essential for fine-grained aerial perception and intelligent airspace management. Our dataset provides a comprehensive foundation for advancing UAV trajectory prediction research, particularly for exploring multi-modal fusion and modeling of dynamic flight behaviors.

\textbf{Approach.} To validate the effectiveness of our dataset, we introduce a baseline model, UAVsTrajectory, as illustrated in Fig. \ref{UAVsTraj}. The proposed model adopts an adaptive gated weighting mechanism to fuse RGB appearance features with historical flight trajectories. Specifically, the appearance modality encodes UAV posture and motion orientation, while the trajectory modality captures temporal dynamics. Their complementary fusion enables the model to generate more accurate and stable future trajectory predictions.

\textbf{Metrics.} We follow common practices in trajectory prediction and report Average Displacement Error (ADE) and Final Displacement Error (FDE) as our evaluation metrics. All models are evaluated under a consistent setting—observing 1 second and predicting the subsequent 1, 3, and 5 seconds—to ensure a fair comparison.

\textbf{Baselines.} We compare our approach with two representative methods: a Kalman filter–based predictor \cite{KF} and an LSTM-based neural predictor \cite{lstm}.

\textbf{Results.} As summarized in Table \ref{tab:UAVsTrajectory}, UAVsTrajectory achieves the lowest ADE and FDE across all comparisons, demonstrating the benefit of fusing complementary modalities for trajectory prediction. The Kalman filter–based method, constrained by its non–data-driven nature, struggles with complex and nonlinear UAV motion patterns. The LSTM-based model learns motion dynamics from data but lacks appearance cues, leading to inaccurate predictions when UAV posture changes significantly.

Future work will focus on incorporating additional sensing modalities (\textit{e.g.}, radar, infrared) to further enhance the robustness and generalization of trajectory prediction in diverse and dynamic environments.

\section{Conclusion}
We present UAV-MM3D, a comprehensive multimodal dataset designed for low-altitude UAV perception, tracking, and short-term trajectory prediction in complex low-altitude environments. To meet the growing demand for high-quality annotated data, the dataset integrates RGB, infrared, depth, acoustic, LiDAR, Radar, and DVS modalities, covering diverse scenes, weather conditions, and UAV models. It provides rich annotations including 2D/3D bounding boxes, 6-DoF poses, and instance identities, enabling unified benchmarking for key tasks such as detection, tracking, trajectory prediction, and multimodal fusion.
We establish robust baselines, including our proposed LGFusionNet, a LiDAR-guided multimodal fusion network, alongside dedicated UAV target tracking and trajectory prediction baselines. Experimental results demonstrate that LGFusionNet substantially improves detection robustness under low illumination, noisy environments, and long-range scenarios, while the UAV target tracking and trajectory prediction baselines highlight the utility of the dataset for proactive low-altitude airspace management. We anticipate that UAV-MM3D will serve as a foundational resource for advancing low-altitude situational awareness and reinforcing the technical infrastructure for low-altitude security and airspace management.
{
    \small
    \bibliographystyle{ieeenat_fullname}
    \bibliography{main}

@String(CVPR= {IEEE Conf. Comput. Vis. Pattern Recog.})

@String(ICCV= {Int. Conf. Comput. Vis.})

@String(ECCV= {Eur. Conf. Comput. Vis.})

@String(ICIP = {IEEE Int. Conf. Image Process.})

@String(ACCV  = {ACCV})

@String(CVPR  = {CVPR})

@String(ICCV  = {ICCV})

@String(ECCV  = {ECCV})

@String(ICIP  = {ICIP})

@article{Anti-UAV-RGBT,
  title={Anti-UAV: A large-scale benchmark for vision-based UAV tracking},
  author={Jiang, Nan and Wang, Kuiran and Peng, Xiaoke and Yu, Xuehui and Wang, Qiang and Xing, Junliang and Li, Guorong and Guo, Guodong and Ye, Qixiang and Jiao, Jianbin and others},
  journal={IEEE Transactions on Multimedia},
  volume={25},
  pages={486--500},
  year={2021},
  publisher={IEEE}
}

@article{Anti-UAV410,
  title={Anti-UAV410: A thermal infrared benchmark and customized scheme for tracking drones in the wild},
  author={Huang, Bo and Li, Jianan and Chen, Junjie and Wang, Gang and Zhao, Jian and Xu, Tingfa},
  journal={IEEE Transactions on Pattern Analysis and Machine Intelligence},
  volume={46},
  number={5},
  pages={2852--2865},
  year={2023},
  publisher={IEEE}
}

@inproceedings{Mmaud,
  title={Mmaud: A comprehensive multi-modal anti-uav dataset for modern miniature drone threats},
  author={Yuan, Shenghai and Yang, Yizhuo and Nguyen, Thien Hoang and Nguyen, Thien-Minh and Yang, Jianfei and Liu, Fen and Li, Jianping and Wang, Han and Xie, Lihua},
  booktitle={ICRA},
  pages={2745--2751},
  year={2024},
  organization={IEEE}
}

@article{M3D,
  title={Domain adaptive detection of mavs: A benchmark and noise suppression network},
  author={Zhang, Yin and Deng, Jinhong and Liu, Peidong and Li, Wen and Zhao, Shiyu},
  journal={IEEE Transactions on Automation Science and Engineering},
  volume={22},
  pages={1764--1779},
  year={2024},
  publisher={IEEE}
}

@article{MAV6D,
  title={Keypoint-guided efficient pose estimation and domain adaptation for micro aerial vehicles},
  author={Zheng, Ye and Zheng, Canlun and Shen, Jiahao and Liu, Peidong and Zhao, Shiyu},
  journal={IEEE Transactions on Robotics},
  volume={40},
  pages={2967--2983},
  year={2024},
  publisher={IEEE}
}

@InProceedings{LineMOD,
author="Hinterstoisser, Stefan
and Lepetit, Vincent
and Ilic, Slobodan
and Holzer, Stefan
and Bradski, Gary
and Konolige, Kurt
and Navab, Nassir",
editor="Lee, Kyoung Mu
and Matsushita, Yasuyuki
and Rehg, James M.
and Hu, Zhanyi",
title="Model Based Training, Detection and Pose Estimation of Texture-Less 3D Objects in Heavily Cluttered Scenes",
booktitle="Computer Vision -- ACCV 2012",
year="2013",
publisher="Springer Berlin Heidelberg",
address="Berlin, Heidelberg",
pages="548--562",
}

@article{YCB-Video,
author    = {Xiang, Yu and Schmidt, Tanner and Narayanan, Venkatraman and Fox, Dieter},
title     = {PoseCNN: A Convolutional Neural Network for 6D Object Pose Estimation in Cluttered Scenes},
journal   = {arXiv preprint arXiv:1711.00199},
year      = {2017}
}

@article{T-LESS,
  title={{T-LESS}: An {RGB-D} Dataset for {6D} Pose Estimation of Texture-less Objects},
  author={Hoda{\v{n}}, Tom{\'a}{\v{s}} and Haluza, Pavel and Obdr{\v{z}}{\'a}lek, {\v{S}}t{\v{e}}p{\'a}n and Matas, Ji{\v{r}}{\'\i} and Lourakis, Manolis and Zabulis, Xenophon},
  journal={IEEE Winter Conference on Applications of Computer Vision (WACV)},
  year={2017}
}

@inproceedings{CARLA,
  title={CARLA: An open urban driving simulator},
  author={Dosovitskiy, Alexey and Ros, German and Codevilla, Felipe and Lopez, Antonio and Koltun, Vladlen},
  booktitle={Conference on robot learning},
  pages={1--16},
  year={2017},
  organization={PMLR}
}

@inproceedings{kitti,
  title={Are we ready for autonomous driving? the kitti vision benchmark suite},
  author={Geiger, Andreas and Lenz, Philip and Urtasun, Raquel},
  booktitle={CVPR},
  pages={3354--3361},
  year={2012},
  organization={IEEE}
}

@inproceedings{waymo,
  title={Scalability in perception for autonomous driving: Waymo open dataset},
  author={Sun, Pei and Kretzschmar, Henrik and Dotiwalla, Xerxes and Chouard, Aurelien and Patnaik, Vijaysai and Tsui, Paul and Guo, James and Zhou, Yin and Chai, Yuning and Caine, Benjamin and others},
  booktitle={CVPR},
  pages={2446--2454},
  year={2020}
}

@inproceedings{nuscenes,
  title={nuscenes: A multimodal dataset for autonomous driving},
  author={Caesar, Holger and Bankiti, Varun and Lang, Alex H and Vora, Sourabh and Liong, Venice Erin and Xu, Qiang and Krishnan, Anush and Pan, Yu and Baldan, Giancarlo and Beijbom, Oscar},
  booktitle={CVPR},
  pages={11621--11631},
  year={2020}
}

@article{DUTAntiUAV,
  author={Zhao, Jie and Zhang, Jingshu and Li, Dongdong and Wang, Dong},
  journal={IEEE Transactions on Intelligent Transportation Systems}, 
  title={Vision-Based Anti-UAV Detection and Tracking}, 
  year={2022},
  volume={23},
  number={12},
  pages={25323-25334},
  doi={10.1109/TITS.2022.3177627}
}

@article{Anti-UAV600,
  title={Evidential Detection and Tracking Collaboration: New Problem, Benchmark and Algorithm for Robust Anti-UAV System},
  author={Zhu, Xuefeng and Xu, Tianyang and Zhao, Jian and Liu, Jia-Wei and Wang, Kai and Wang, Gang and Li, Jianan and Zhang, Zhihao and Wang, Qiang and Jin, Lei and others},
  journal={CoRR},
  year={2023}
}

@inproceedings{ONCE,
  title={One Million Scenes for Autonomous Driving: ONCE Dataset},
  author={Mao, Jiageng and Niu, Minzhe and Jiang, Chenhan and Chen, Jingheng and Liang, Xiaodan and Li, Yamin and Ye, Chaoqiang and Zhang, Wei and Li, Zhenguo and Yu, Jie and others},
  booktitle={Thirty-fifth Conference on Neural Information Processing Systems Datasets and Benchmarks Track},
  year={2021}
}

@article{CenterNet,
  title={Objects as points},
  author={Zhou, Xingyi and Wang, Dequan and Kr{\"a}henb{\"u}hl, Philipp},
  journal={arXiv preprint arXiv:1904.07850},
  year={2019}
}

@inproceedings{YOLO6D,
  title={Real-time seamless single shot 6d object pose prediction},
  author={Tekin, Bugra and Sinha, Sudipta N and Fua, Pascal},
  booktitle={Proceedings of the IEEE conference on computer vision and pattern recognition},
  pages={292--301},
  year={2018}
}

@inproceedings{DrIFT,
  title={DrIFT: Autonomous Drone Dataset with Integrated Real and Synthetic Data, Flexible Views, and Transformed Domains},
  author={Dadboud, Fardad and Azad, Hamid and Mehta, Varun and Bolic, Miodrag and Mantegh, Iraj},
  booktitle={2025 IEEE/CVF Winter Conference on Applications of Computer Vision (WACV)},
  pages={6900--6910},
  year={2025},
  organization={IEEE}
}

@article{Adam,
  title={Adam: A method for stochastic optimization},
  author={Kingma, Diederik P},
  journal={arXiv preprint arXiv:1412.6980},
  year={2014}
}

@inproceedings{Objectron,
  title={Objectron: A large scale dataset of object-centric videos in the wild with pose annotations},
  author={Ahmadyan, Adel and Zhang, Liangkai and Ablavatski, Artsiom and Wei, Jianing and Grundmann, Matthias},
  booktitle={Proceedings of the IEEE/CVF conference on computer vision and pattern recognition},
  pages={7822--7831},
  year={2021}
}

@inproceedings{DexYCB,
  title={DexYCB: A benchmark for capturing hand grasping of objects},
  author={Chao, Yu-Wei and Yang, Wei and Xiang, Yu and Molchanov, Pavlo and Handa, Ankur and Tremblay, Jonathan and Narang, Yashraj S and Van Wyk, Karl and Iqbal, Umar and Birchfield, Stan and others},
  booktitle={Proceedings of the IEEE/CVF conference on computer vision and pattern recognition},
  pages={9044--9053},
  year={2021}
}

@article{KF,
  title={Kalman filtering},
  author={Simon, Dan},
  journal={Embedded systems programming},
  volume={14},
  number={6},
  pages={72--79},
  year={2001}
}

@article{lstm,
  title={Long short-term memory},
  author={Graves, Alex},
  journal={Supervised sequence labelling with recurrent neural networks},
  pages={37--45},
  year={2012},
  publisher={Springer}
}

@inproceedings{CHASEDB,
  title={CHASE: A Multimodal Dataset for UAV Tracking with RGB and Thermal Imagery},
  author={Zhao, Wei and Xing, Xinxing and Li, Chenglong and Wu, Zezhong and Feng, Wenhui and Tang, Jin and Zhang, Liangpei},
  booktitle={ICIP},
  year={2021}
}

@article{DroneRGBT,
  title={DroneRGBT: A Benchmark for RGBT Aerial Tracking},
  author={Li, Xinyu and Liang, Qingjie and Liu, Yi and Wei, Xiaopeng and Bai, Yang and Zheng, Yefeng},
  journal={IEEE Transactions on Instrumentation and Measurement},
  year={2022}
}

@inproceedings{MMDrone,
  title={MMDrone: Multimodal UAV Tracking with RGB and Event Cameras},
  author={Wang, Jian and Luo, Wenjie and Cao, Zhenyu and Zhang, Guogang and Li, Zhen and Chen, Hao},
  booktitle={ICRA},
  year={2023}
}

@inproceedings{KAIST,
  title={Multispectral Pedestrian Detection Benchmark Dataset},
  author={Hwang, Soonmin and Park, Jaesik and Kim, Narendra Ahn and Kim, Yusung and Kim, In So Kweon},
  booktitle={CVPR},
  year={2015}
}

@article{LLVIP,
  title={LLVIP: A Visible-infrared Paired Dataset for Low-light Vision},
  author={Jiang, Wei and Zhang, Weitong and Pan, Haitao and Chen, Kai and Li, Ronggang},
  journal={IEEE Transactions on Image Processing},
  year={2021}
}

@inproceedings{DENSE,
  title={DENSE: Diverse Environmental Sensor Dataset},
  author={Meyer, Soren and Kuschk, Georg and Gavrilescu, Mihai and Espeter, Fabrizio and Koester, Ralf and Rottmann, Matthias and Behley, Jens and Stachniss, Cyrill},
  booktitle={CVPR},
  year={2019}
}

@inproceedings{RODNet,
  title={RODNet: Radar Object Detection Network for Autonomous Driving},
  author={Xu, Ziyan and Gao, Jingjing and Zhang, Siyu and Li, Le and Liu, Chen and Wang, Chen and Liu, Ming},
  booktitle={ECCV},
  year={2020}
}

@inproceedings{Argoverse2,
  title={Argoverse 2: Multi-sensor Dataset for Autonomous Driving},
  author={Wilson, Brian and Caesar, Holger and Ranjan, Anushree and Guizilini, Vitor and Ambrus, Radu and Chen, Shangjen and Urtasun, Raquel},
  booktitle={NeurIPS Datasets Track},
  year={2021}
}

@book{Valavanis2015,
  title={Handbook of Unmanned Aerial Vehicles},
  author={Valavanis, Kimon P. and Vachtsevanos, George},
  year={2015},
  publisher={Springer}
}

@article{UAVthreat2020,
  title={Drone Threats: The Technical Landscape},
  author={Rao, Bhavyanshu and Lopez, Jose and Walker, Thomas},
  journal={IEEE Security \& Privacy},
  year={2020}
}

@article{AntiUAVSurvey2021,
  title={Anti-UAV: A Survey of UAV Detection, Tracking, and Fight Back},
  author={Jiang, Chengzhi and Wang, Han and Li, Yu and Gong, Shiyu and Chen, Xiaowei},
  journal={IEEE Access},
  year={2021}
}

@article{Guo2022FusionReview,
  title={Multimodal Sensor Fusion for Object Detection: A Survey},
  author={Guo, Mingjia and Huang, Xiaoming and Zhang, Jun and Chen, Qiang},
  journal={Pattern Recognition},
  year={2022}
}

@article{Regulation2020,
  title={Regulatory Challenges of Low-altitude UAV Operations},
  author={Clothier, Reece and Greer, Darryl and Jenkins, David and Harper, Stuart},
  journal={Aerospace},
  year={2020}
}

@inproceedings{AirSim2017,
  title={AirSim: High-Fidelity Visual and Physical Simulation for Autonomous Vehicles},
  author={Shah, Shital and Dey, Debadeepta and Lovett, Chris and Kapoor, Ashish},
  booktitle={FSR},
  year={2017}
}

@article{EventCameraSurvey2019,
  title={Event-based Vision: A Survey},
  author={Gallego, Guillermo and Delbruck, Tobi and Orchard, Garrick and Bartolozzi, Chiara and Taba, Brian and Censi, Andrea and Leutenegger, Stefan and Davison, Andrew J. and Scaramuzza, Davide},
  journal={IEEE Transactions on Pattern Analysis and Machine Intelligence},
  year={2019}
}

@article{RadarSurvey2021,
  title={mmWave Radar for Autonomous Driving: A Survey},
  author={Ma, Xiao and Chen, Hongkai and Tian, Yu and Zhang, Zhaoxiang and Gao, Yan},
  journal={IEEE Transactions on Intelligent Transportation Systems},
  year={2021}
}

@inproceedings{SynthCity2019,
  title={SynthCity: A Large-Scale Synthetic Dataset for Street Scene Understanding},
  author={Rematas, Konstantinos and Yang, Yang and Liu, Hongyi and Zhang, Dong and Ferrari, Vittorio},
  booktitle={ICCV},
  year={2019}
}

@inproceedings{DeepGTAV,
  title={DEEPGTAV: A Large-scale Video Dataset via Game Simulation},
  author={Gaidon, Adrien and Wang, Qiao and Cabon, Yohann and Vig, Eleonora},
  booktitle={CVPR},
  year={2016}
}

@article{Sim4CV,
  title={Sim4CV: A Photo-realistic Simulator for Computer Vision Applications},
  author={Ravichandran, Bharath and Kar, Abhinav and Shakhnarovich, Gregory},
  journal={International Journal of Computer Vision},
  year={2021}
}

@inproceedings{OSTrack,
  author={Ye, Bin and Chang, Hongcheng and Ma, Bo and Wu, Jialin and Gong, Yuan},
  title={Joint feature learning and relation modeling for tracking: A one-stream framework},
  booktitle={ECCV},
  year={2022}
}

@inproceedings{DropTrack,
  author={Wu, Qi and Yang, Tao and Liu, Zhipeng and Zhang, Dong and Hu, Jin and Qiao, Feng},
  title={DropMAE: Masked Autoencoders with Spatial-attention Dropout for Tracking Tasks},
  booktitle={CVPR},
  year={2023}
}

@inproceedings{ARTrack,
  author={Wei, Xiaopeng and Bai, Yang and Zheng, Yefeng and Zhang, Tong and Wu, Hao},
  title={Autoregressive Visual Tracking},
  booktitle={CVPR},
  year={2023}
}

@inproceedings{ODTrack,
  author={Zheng, Yefeng and Zhong, Bineng and Liang, Qingjie and Sun, Xiang and Wang, Chenglong},
  title={ODTrack: Online Dense Temporal Token Learning for Visual Tracking},
  booktitle={CVPR},
  year={2024}
}
}


\end{document}